\documentclass[conference]{IEEEtran}
\usepackage{xcolor}
\usepackage{bm}
\usepackage{bbm}
\usepackage{amsfonts}
\usepackage{mathrsfs}
\usepackage{amssymb}
\usepackage{graphicx}
\usepackage{amsmath}
\usepackage{subfigure}
\usepackage{hyperref}
\usepackage{float}
\usepackage{array}
\usepackage{caption}

\begin{document}
\title{GANmut: Generating and Modifying Facial Expressions}

\author{
\IEEEauthorblockN{Maria Surani}
\IEEEauthorblockA{Queen Mary\\
University of London\\
Email: ec21676@qmul.ac.uk}

}

\maketitle

\begin{abstract}
In the realm of emotion synthesis, the ability to create authentic and nuanced facial expressions continues to gain importance. Traditional generative adversarial networks (GANs) have been constrained by the specific expressions represented in their training datasets, often limited by exhaustive annotation requirements or a conflated label space. Recognizing these limitations, the GANmut study discusses a recently introduced advanced GAN framework that, instead of relying on predefined labels, learns a dynamic and interpretable emotion space. This methodology maps each discrete emotion as vectors starting from a neutral state, their magnitude reflecting the emotion's intensity.The current project aims to extend the study of this framework by benchmarking across various datasets, image resolutions, and facial detection methodologies. This will involve conducting a series of experiments using two emotional datasets: Aff-Wild2 and AffNet. Aff-Wild2 contains videos captured in uncontrolled environments, which include diverse camera angles, head positions, and lighting conditions, providing a real-world challenge. AffNet offers images with labelled emotions, improving the diversity of emotional expressions available for training. The first two experiments will focus on training GANmut using the Aff-Wild2 dataset, processed with either RetinaFace or MTCNN—both of which are high-performance deep learning face detectors. This setup will help determine how well GANmut can learn to synthesise emotions under challenging conditions and assess the comparative effectiveness of these face detection technologies. The subsequent two experiments will merge the Aff-Wild2 and AffNet datasets, combining the real-world variability of Aff-Wild2 with the diverse emotional labels of AffNet. The same face detectors, RetinaFace and MTCNN, will be employed to evaluate whether the enhanced diversity of the combined datasets improves GANmut's performance and to compare the impact of each face detection method in this hybrid setup.

\end{abstract}

\IEEEpeerreviewmaketitle

\section{Introduction}

Facial expressions are fundamental components of non-verbal communication, crucial for conveying a broad range of human emotions. Extensive research has explored the relationship between expressions and emotions \cite{russell1977evidence, russell1980circumplex, kanade2000comprehensive}, encountering challenges such as the ambiguity of the same expression representing different emotions; the difficulty in distinguishing closely related emotions, which are separated only by subtle variations in facial movements; and inconsistencies in labelling due to subjectivity concerns. Furthermore, cultural backgrounds influence how facial expressions are interpreted and used.

In psychology, numerous models of emotion recognition have been proposed, with researchers striving to achieve a unified understanding. However, over the past decade, machine learning has emerged as a promising alternative for modelling emotional expressions from a computational perspective. Machine learning methods offer the advantage of reduced bias compared to traditional methods, which often suffer from the limitations of observations and inherent personal biases of human analysts like psychologists \cite{kollias2023deep2,arsenos2023data}. Machine learning algorithms and computational techniques introduce minimal prior context and can analyse significantly larger datasets before modelling new emotional instances. However, maintaining the interpretability of these emotions for human understanding remains a challenge.

In exploring the interpretability of emotions within current research, several challenges arise. If the goal is to model straightforward, laboratory-controlled expressions, a category-based split of emotions might suffice \cite{larsen1992promises, ekman1971constants}. However, the problem becomes more complex with spontaneous expressions, where ambiguities are increased due to non-expert consensus and labelling discrepancies. Relying solely on these labels for data collection is largely ineffective. More complex models might be considered at the expense of simplicity, such as compound emotions \cite{du2014compound}, Valance-Arousal (VA) \cite{russell1977evidence}, and Action Units (AUs) \cite{ekman2002facial}. 

\begin{figure}[h]
    \centering
    \includegraphics[width=0.95\columnwidth]{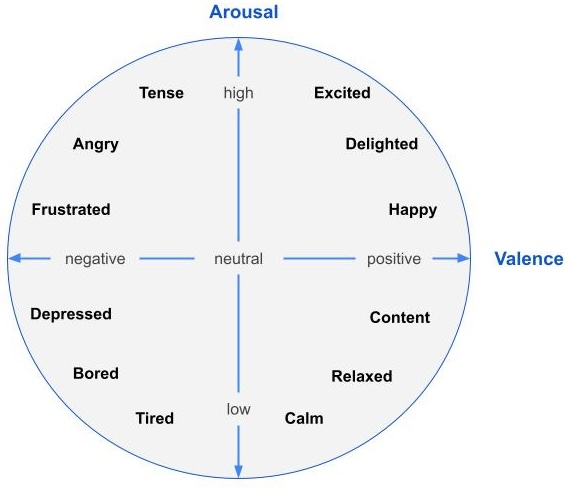}
    \caption{Valence-Arousal 2D space}
    \label{fig:enter-label}
\end{figure}

\begin{figure}[h]
    \centering
    \includegraphics[width=0.95\columnwidth]{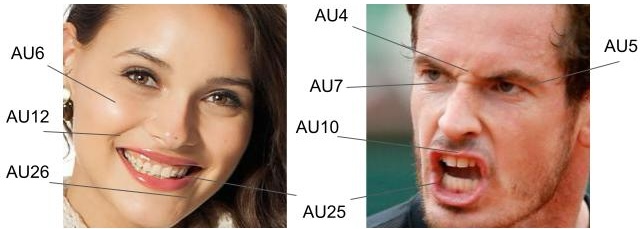}
    \caption{Visualization of the facial action units (AUs) used to express happiness (left) and anger (right). AU4: brow lowerer, AU5: upper lid raiser, AU6: cheek raiser, AU7: lid tightener, AU10: upper lip raiser, AU12: lip corner puller, AU25: lip part, AU26: jaw drop \cite{affnet}}
    \label{fig:enter-label}
\end{figure}

A promising approach could involve developing a learning algorithm capable of utilising imperfect emotion labels. In terms of machine learning, advancing further would mean learning the entire spectrum of emotional labels and then mapping these to the imperfect labels available.

\begin{figure}[h]
    \centering
    \includegraphics[width=0.95\columnwidth]{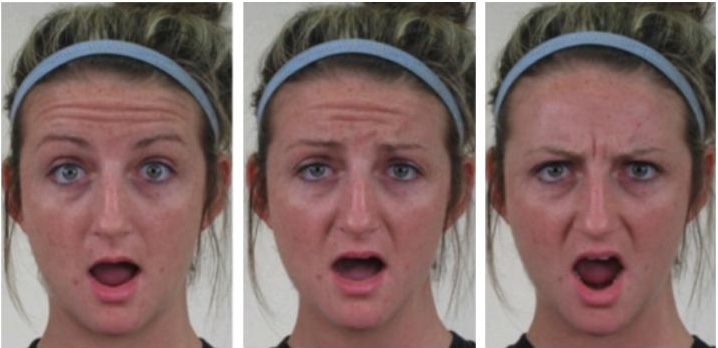}
    \caption{Left - right: surprised, sadly surprised, angrily surprised \cite{Du2014CompoundFE}}
    \label{fig:compound_emotions}
\end{figure}

A persistent issue in labelling is the attempt to discretize a continuous emotional spectrum, which can be seen through our facial muscles and the nuances of expressions. This discretization process often misses out on the intensity and the fluid transition between emotions. For instance, the expression of surprise may overlap with sadness or anger, creating a complex blend (Fig. \ref{fig:compound_emotions}) . 

By incorporating intensity, the ambiguous nature of emotions becomes more pronounced compared to basic, singular emotions. It becomes essential to recognize intensity measurements and intermediate emotions to make categorical labels more effective. Therefore, categorical labels should be used primarily as heuristics to approximate the underlying emotional states.

\begin{figure}[h]
    \centering
    \includegraphics[width=0.95\columnwidth]{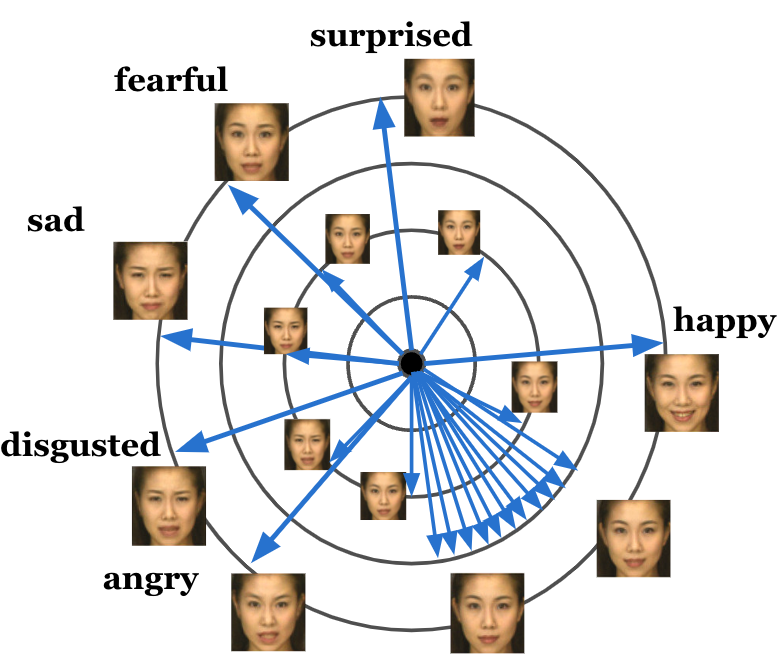}
    \caption{Gamut of Emotions: built using \cite{song2018selective}.}
    \label{fig:enter-label}
\end{figure}

Fig. \ref{fig:enter-label} illustrates each discrete emotion as a vector originating from a neutral central point. The length of each vector indicates the emotion's intensity, while the blend of high-level, basic emotions is illustrated by using samples from the intermediate space that exists between the vectors representing basic emotions. The model samples from the conditional space along these vectors, aiming to preserve the category with confidence based on the intensity. The images produced should not only be realistic but also recognizable within the learned conditional space. Therefore, every image must be realistically generated and capable of being inversely mapped, ensuring that the expressions generated accurately represent the distribution of real images.

The learned conditional space is continuous and interpretable in order to generate the full spectrum of emotions in Fig. \ref{fig:enter-label}. Traditional conditional GANs often fall short in this regard because they employ classification loss, which tends to produce a generator that only creates easily identifiable emotions. In contrast, the GANmut \cite{d2021ganmut} approach enhances complexity by incorporating nuanced labels and their corresponding expressions.

This study aims to further evaluate GANmut’s approach to generating emotions and improve the accuracy and depth of its quantitative and qualitative assessments. The key contributions of this project include: introducing new data sources featuring subjects in uncontrolled environments with varying lighting conditions and head positions, which add layers of complexity; expanding performance analysis to include additional facial detection algorithms; and conducting more quantitative evaluations. 

\section{Related Work}

\subsection{Deep Learning and Generative Adversarial Networks}

Deep learning as a framework aims to discover hierarchical models representing probability distributions over the types of data that typically appears in artificial intelligence such as natural images, audio waveforms, speech patterns and symbols in natural language form \cite{bengio2009learning,kollias2023ai,kollias2023ai,kollias2023btdnet,gerogiannis2024covid}. The starting point has been discriminative models mapping high dimensional, sensory inputs to a class label \cite{hinton2012deep,salpea2022medical}. The ample successes in this area are based on backpropagation and the dropout algorithm, using piecewise linear units with well-behaved gradients \cite{glorot2011deep}. In contrast, Deep generative models have had less impact due to difficult approximations of many intractable probabilistic computations that arise in maximum likelihood estimation (MLE) and similar strategies, and generally due to difficulties in using piecewise linear units in generative contexts.

The current work is part of a larger class of algorithms that leverage Generative Adversarial Networks (GANs) \cite{goodfellow2014generative}. A GAN consists of two primary components: a generator ($G$) and a discriminator ($D$). The generator is a directed latent variable model that deterministically creates samples $x$ from noise vectors $z$. The discriminator, on the other hand, aims to distinguish between samples drawn from the actual dataset and those produced by the generator. This setup creates a dynamic training environment characterised by a mini-max game, in which the generator and discriminator continuously improve by reacting to each other's progress. 

The adversarial modeling framework has the most straightforward applications when the models are both multi-layer perceptrons. We learn the generator's distribution $p_g$ over data $\bm{x}$ by defining a prior on input noise variables $p_{\bm{z}}(\bm{z})$, then represent a mapping to data space as $G(\bm{z}; \theta_g)$, for $G$ a differentiable function represented by a multi-layer perceptron parameterized by $\theta_g$. A second multilayer perceptron $D(\bm{x}; \theta_d)$ is defined that outputs a single scalar. We train $D$ to maximize the probability of assigning the correct label to both training examples and samples from $G$, while simultaneously training $G$ to generate data indistinguishable from the real one. $D$ and $G$ play the following two-player mini-max game with value function $V(G, D)$:

\begin{align}
\begin{array}{lll}
    & \min _G \max _D V(D, G) = &\mathbb{E}_{\boldsymbol{x} \sim p_{\text {data }}(\boldsymbol{x})}[\log D(\boldsymbol{x})] + \\
     & & \mathbb{E}_{\boldsymbol{z} \sim p_{\boldsymbol{z}}(\boldsymbol{z})}[\log (1-D(G(\boldsymbol{z})))]
\end{array}
\end{align}

\subsection{Facial Expressions: Label Conditioning}
Facial expression and emotion detection has been researched actively in computer vision for the past decades. The field of affective computing particularly references three main models: 1) Basic emotions are described in \cite{ekman1971constants} which uses seven discrete emotions extending the ones in Fig. \ref{fig:enter-label} with "interest". 2) Valence-Arousal (VA) model \cite{russell1977evidence} which maps facial emotions in a continuous 2D space with Valence (i.e., positive / negative scale) and Arousal (i.e., intensity of emotion) as coordinates. 3) Action Units (AUs) provide an even finer definition in terms of facial muscle movements \cite{ekman2002facial}, modelling expressions as contractions / relaxations of various muscles. Emotion annotation is therefore based on the above mentioned models. Categorical models are preferred allowing for non-expert labelling in the detriment of less representative spaces. Although VA and AUs can represent much larger sets of emotions, VA introduces ambiguity through only using two coordinates (e.g., scared and angry expressions have both high arousal and low valence), while AUs demand expensive labelling processes and do not reveal emotion directly.

Existing methods that use GANs \cite{goodfellow2020generative} produce good results in settings with categorical emotions, with major shortcomings in extending beyond labelled emotion definition. StarGAN \cite{choi2018stargan} uses conditional GANs to push categorical labels in the generator and produce domain specific images. GANimation \cite{pumarola2018ganimation} uses AU conditioning exploiting a similar scheme with focus on local transformations using the attention framework. SMIT \cite{romero2019smit} switches the StarGANs deterministic output into stochastic noise manipulation producing many outputs from a single input. To address their inability to extend beyond categorical emotions, StarGAN-v2 \cite{choi2020stargan} is an alternative that produces realistic conditioned image manipulations through interpolations. However, as noted in \cite{romero2021smile}, StarGAN-v2 fails for visually close domains such as spontaneous facial attributes and emotions.

Conditional GANs \cite{odena2017conditional} have been used in prior papers as an alternative by assuming auxiliary classifiers that encourage the generator to produce well-defined labels across the same set of classes. This strategy, however, tends to reinforce only the predefined categorical labels, limiting the representation of nuanced or spontaneous expressions, such as "happily surprised" or "sadly fearful". Therefore, conditioning on basic emotions alone does not suffice to encompass a wide emotional range.

The GANmut approach aims to learn an expressive conditional space $Z$ in which producing a gamut of potentially unlabelled emotions becomes possible \ref{fig:enter-label}.

\subsection{GANs Latent Spaces}

The following subsection will relate to various ways of capturing the conditional space and injecting additional information on the generator side and still allowing for manipulations. The main problem of finding interpretable directions in the latent space or input that can be mapped further to changes in a generated image (e.g., happiness, age appearance, or any other aspects) by the GAN's generator was studied in \cite{voynov2020unsupervised}, where the authors propose a method based on a reconstructor. Their method aims to learn a set of $n$ independent, recognizable directions based on a uniformly sampled $\epsilon \sim \mathcal{U}[-c, c]$. Other relevant works in the area of semantic structure of GAN latent spaces include \cite{harkonen2020ganspace}, \cite{shen2020interpreting}. Another interesting problem in this area comes from interpolating two latent codes in order to generate gradual transitions between images. Unfortunately, the linear path might not always lead to smooth transitions. A method has been proposed \cite{laine2018feature} to search for making these straight paths meaningful by minimizing a feature-based loss function, which was also used in the GANmut paper.

\section{Problem Formulation}

The focus of GANmut, \cite{d2021ganmut}, is to produce arbitrary emotional facial expressions for a continuous space, given a dataset of real facial expression images $X=\left\{x_{1}, \ldots, x_{N}\right\}$ and their categorical labels $C=\left\{c_{1}, \ldots, c_{N}\right\}, c_{i} \in\{1, \ldots, M\}$, where $M$ is usually 7. The solution proposed in \cite{d2021ganmut} is to exploit the methodology of conditional generative adversarial networks (GANs). This framework assumes an adversarial game between a generator $G$ and a discriminator $D$, that approximates the distribution of the real facial expressions conditioned on the labels. The main issue is that basic emotion labels cannot cover the Gamut of emotions of real facial expressions due to ambiguity described in previous sections, with multi-emotion states, for example happily disgusted being labelled just as happy. At the same time, the conventional GAN methods rely on a static conditional space $Z$ that only encodes the given labels, which is expected to approximate the distributions only over pure basic emotions. Therefore, the problem becomes making the GAN conditional space learnable, which can help with the imperfect emotion labelling by discovering the complete distribution of the given arbitrary facial expressions. 
\cite{sym15040956}

The first approach proposed in \cite{d2021ganmut} is to exploit parameterization techniques on the GAN conditional space to encode the full Gamut. A feasible strategy is to use a two-dimensional conditional space $Z$ defined in polar coordinates $(\theta, \rho)$; where the random variable $z = (\theta, \rho)$ represents the conditional latent space. The coordinates of the space are derived from a uniform distribution $\theta \sim \mathcal{U}([0,2\pi]), \quad \rho \sim \mathcal{U}([0,1])$. In this context, $\theta$ corresponds to the category of emotion, while $\rho$ is associated with its intensity. The condition of each basic emotion would be defined using $z = (\theta, \cdot)$, where $\theta_{c_i}$ represents the learned direction corresponding to emotion $c_i$.

\begin{figure}[H]
    \centering
    \includegraphics[width=0.95\columnwidth]{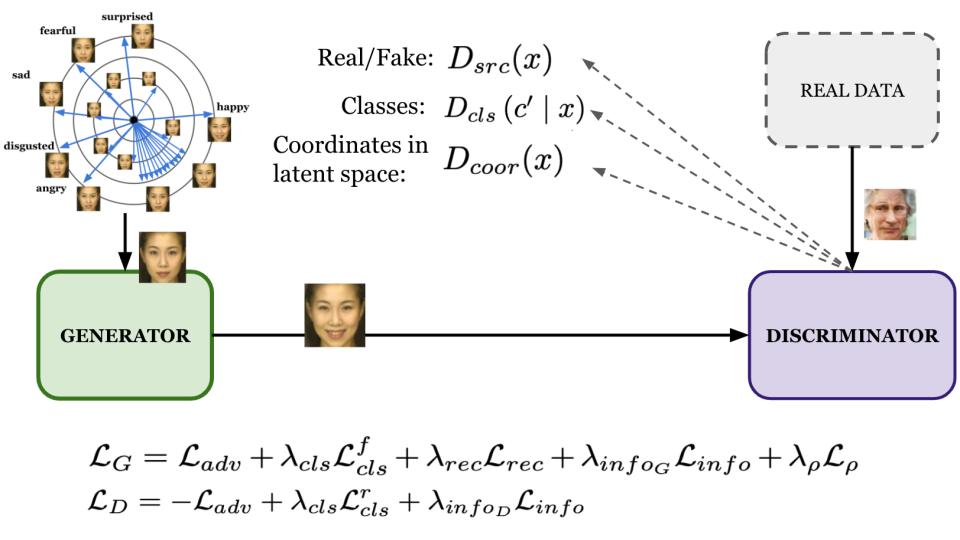}
    \caption{GANmut Architecture and Flow.}
    \label{fig:ganmutarch}
\end{figure}

\section{Methodology}
The approach defined in \cite{d2021ganmut} aims to learn the GANmut conditional space following a key central concept: substituting the existing static conditional space for the parameterized, dynamic space described earlier. This can be achieved by using a standard multi-domain conditional GAN model (such as StarGAN \cite{choi2018stargan}), enabling both the parameterised model and space to be optimized simultaneously through adversarial training techniques. The proposed polar parameterization \cite{d2021ganmut} of the conditional space not only facilitates sampling from labeled basic emotions but also enables exploration within the space's representation. This enhanced and dynamic exploration of the conditional space helps in identifying the emotion spectrum through a combined approach of adversarial, classification, and regression losses. These losses are essential for refining the generator and discriminator, enabling them to generate authentic emotional facial expressions, accurately classify basic emotion labels, and regress the continuous latent variables.

\subsection{Linear Model}
The model introduces a new sampling strategy within a "conditional space" of emotions. The conditions associated with the basic emotions can be progressively optimized, while the conditions associated with more complex emotions are sampled and updated simultaneously. 

For each basic emotion $c_i$, the model samples a sequence of related latent code $\hat{z} = (\theta_{c_i}, \hat{\rho}_{j,ci})$. In this process $\hat{\rho}_{j}$ is gradually increased, meaning that $\hat{z}_j$ moves outward from the origin in the direction of $\theta _{c_i}$, while $\theta _{c_i}$ is updated to ensure that the generated images $y_{\hat{z}_j}$ are increasingly identified as emotion $c_i$. The discriminator $D$ should recognize $y_{\hat{z}}$ as emotion $c_i$, based on the distance $\hat{\rho}_{c_i}$ from the origin. If $\hat{\rho}_{c_i}$ is below a set threshold $\mathcal{T}$ (set at 0.2 for the study), $D$ should classify $y_{\hat{z}}$ as expressing a neutral emotion. For more nuanced or mixed emotions, the model uses a different sampling strategy that can combine characteristics from multiple basic emotions.

The development of the conditional space (the way emotions are represented and varied) and the effectiveness of the GAN are interdependent. A more refined conditional space leads to more realistic emotional expressions, while a better-performing GAN enhances the model's ability to understand and manipulate the emotional space. To effectively develop both components, the model uses a training approach that simultaneously optimises the GAN and its understanding of the emotional space.
The model uses a combination of several loss functions for training:

Standard GAN Loss ($\mathcal{L}_{adv}$): ensures the generated images resemble real data and can fool a discriminator network - combined with Wasserstein Loss \cite{arjovsky2017wasserstein}. 

\begin{align}
\mathcal{L}_{adv} &=\mathbb{E}_x\left[D_{src}(x)\right]-\mathbb{E}_{x, z}\left[D_{src}(G(x, z))\right] 
\end{align}

Emotion Classification Loss ($\mathcal{L}_{cls}$): helps align the generated expressions with the intended emotions based on human-assigned labels.

\begin{align}
    \mathcal{L}_{cls}^f &=\mathbb{E}_{x, c, \rho}\left[-\log D_{cls}\left(c \mid G\left(x, z_{c, \rho}\left(\theta_c\right)\right)\right)\right] \\
    \mathcal{L}_{cls}^r &=\mathbb{E}_{x, c'}\left[-\log D_{cls}\left(c' \mid x\right)\right] 
\end{align}

Condition Regression Loss ($\mathcal{L}_{info}$): encourages the discriminator to accurately estimate the emotional coordinates of both real and generated images.

\begin{align}
    \mathcal{L}_{info} &=\mathbb{E}_{x, z}\left[\| D_{coor }(G(x, z))-z \|_2^2\right] 
\end{align}

Interpolation Loss ($\mathcal{L}_\rho$): ensures that moving outward in the emotional space leads to more distinct and intense expressions.

\begin{align}
    \mathcal{L}_\rho=\mathbb{E}_{x, z}\left[\|\hat{\rho}(G(x, z \cdot, \rho))-\rho\|_2^2 \mathbbm{1}_{\rho>0.2}\right]
\end{align}

$\mathcal{L}_{rec}$ (\cite{choi2018stargan}, \cite{zhu2017unpaired}) helps regularise the model and improve its overall performance.

\begin{align}
    \mathcal{L}_{rec} &=\mathbb{E}_{x, z}\left[\left\|x-G\left(G(x, z), D_{coor }(x)\right)\right\|_1\right]
\end{align}

By combining these losses, the model can simultaneously optimise the GAN itself and its understanding of the emotional space:

\begin{align}
    \mathcal{L}_D &=-\mathcal{L}_{adv}+\lambda_{cls} \mathcal{L}_{cls}^r+\lambda_{info_D} \mathcal{L}_{info} \\
    & \quad -\lambda_{gp} \mathbb{E}_{\hat{x}}\left[\left(\left\|\nabla_{\hat{x}} D_{src}(\hat{x})\right\|_2-1\right)^2\right]  \nonumber \\
    \mathcal{L}_G &=\mathcal{L}_{adv}+\lambda_{cls} \mathcal{L}_{cls}^f+\lambda_{rec} \mathcal{L}_{rec}+\lambda_{info_G} \mathcal{L}_{info}+\lambda_\rho \mathcal{L}_\rho 
\end{align}

\section{Face Detectors}
The DeepFace library \cite{deepface} is an open-source Python framework that simplifies facial analysis tasks by integrating multiple face detectors, including high-accuracy models like RetinaFace and MTCNN. While it offers ease of installation and abstracts complex model details, it limits customization due to its lack of parameter fine-tuning flexibility.

\begin{figure}[H]
    \centering
    \includegraphics[width=1\columnwidth]{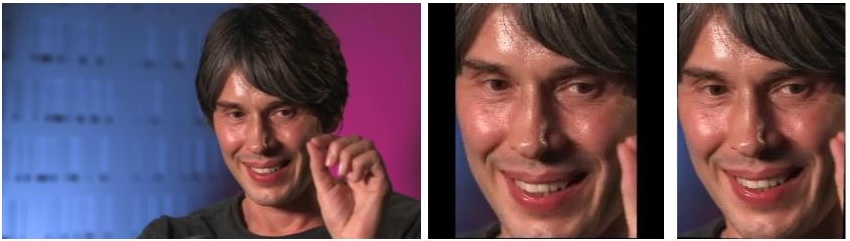}
    \caption{Left - right: Aff-Wild2 original, MTCNN, RetinaFace.}
    \label{fig:face_detectors}
\end{figure}

\subsection{RetinaFace}
RetinaFace \cite{retinaface} employs a single-stage deep learning architecture that integrates face detection and landmark identification into one unified model. It uses a pre-trained convolutional neural network (CNN) to extract features from images, identifying potential facial regions. To address the challenge of detecting faces at various scales, it incorporates a Feature Pyramid Network to create a multi-scale pyramid of feature maps. This approach ensures accurate representation of faces of different sizes.

RetinaFace's multi-task learning strategy performs face classification, bounding box regression, and landmark prediction simultaneously across different levels of the pyramid. This comprehensive method enhances its ability to pinpoint and confirm the presence of faces by accurately locating key facial landmarks such as eyes, nose, and mouth corners.

However, RetinaFace faces challenges with the AffectNet dataset’s pre-cropped images, often resulting in missed detections and false positives due to incomplete bounding boxes for faces positioned very close to the edges.

\subsection{MTCNN}
MTCNN \cite{mtcnn} operates through a cascaded structure involving three stages of deep convolutional networks (CNNs) that refine face detection and landmark localization in a coarse-to-fine manner. Initially, it scans the entire image at low resolution to identify potential face regions, producing bounding boxes with associated probability scores. Subsequent stages focus on the most promising areas, enhancing the precision of the boxes and the confidence in these detections by discarding weaker candidates and refining the positioning at progressively higher resolutions. This cascaded approach not only ensures computational efficiency by quickly eliminating non-face regions but also minimizes false positives, resulting in highly accurate face detection and landmark prediction. In contexts like the cropped AffectNet dataset, MTCNN's flexibility and robust design allow it to handle incomplete information effectively, providing better detection outcomes for edge faces than some alternatives like RetinaFace.

\subsection{Data Sources}

The original GANmut study used the full-size AffectNet \cite{affnet} dataset, containing about 1 million images gathered from queries on popular search engines using 1250 keywords in six languages, representing seven fundamental emotions. For benchmarking, the Aff-Wild2 \cite{kollias2019deep,kollias2019expression,kollias2022abaw,kollias2023abaww,kollias2023btdnet,kollias2023facernet,kollias2023multi,kollias20246th,kollias2024distribution,kollias2024domain,kolliasijcv,zafeiriou2017aff,hu2024bridging,psaroudakis2022mixaugment,kollias2020analysing,kollias2021distribution,kollias2021affect,kollias2019face,kollias2021analysing} dataset, featuring natural, unposed videos of varied resolutions and conditions \cite{kollias2023abaw2, kollias2022abaw, kollias2021affect}, was used to evaluate the model’s ability to recognize nuanced human emotions in challenging environments. The model’s testing was further enhanced by integrating Aff-Wild2 with the small-version of AffectNet (cropped to portray only the face), combining the real-world emotional transitions of Aff-Wild2 \cite{kollias2021distribution, kollias2020analysing, kollias2021analysing} with the extensive diversity of AffectNet. This mixed dataset aims to simulate the complex nature of human emotions in a wide array of facial expressions and scenarios. Data preparation for Aff-Wild2 involved breaking down videos into frames according to their frames-per-second (FPS) rate, with each frame labeled for the emotion depicted. Both face detectors return a list of identified faces, sorted by the confidence score. The first returned face was used for training, following the assumption that the face with the highest confidence score is the one that was annotated. 

\begin{figure}[h]
    \centering
    \includegraphics[width=0.95\columnwidth]{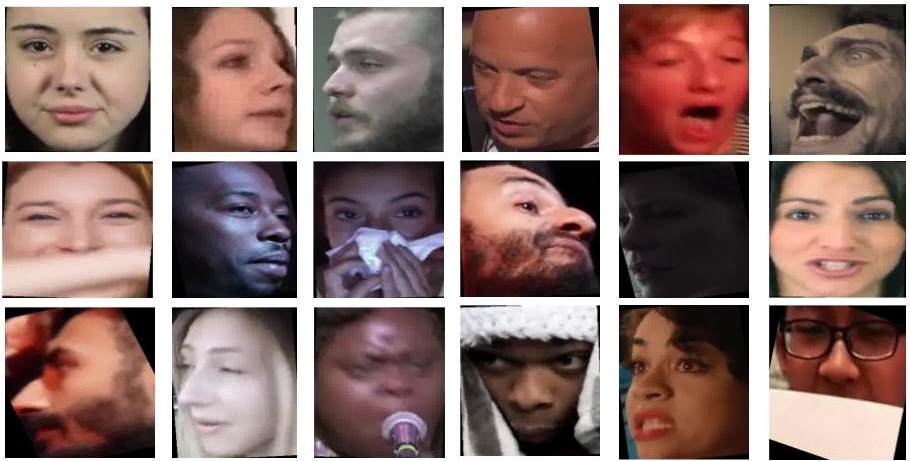}
    \caption{Examples of difficult conditions in the Aff-Wild2 Dataset}
    \label{fig:Aff-Wild2_samples}
\end{figure}

Despite initially considering the RAF-DB \cite{rafdb} dataset, its uneven emotion distribution led to the selection of AffectNet as the more effective training resource for the classifiers used to compute the numerical evaluation.

\subsection{Training}
During training, GANmut employs a custom data loader that preprocesses images for training by reading a CSV file containing image paths and corresponding emotional labels. The data loader standardizes images to 128 x 128 pixels, normalizes the data, and applies data augmentation techniques like random rotations, translations, and zoom to enhance model generalization.

\textbf{Building the CSV training file}

The CSV file is built from the Aff-Wild2 training dataset, which consists of 245 videos. Frames are extracted using ffmpeg, organized by their frames-per-second (fps) rate, and stored in dedicated folders. The script then correlates each frame with emotion labels from annotation files, discarding irrelevant frames and remapping labels to match those used in AffectNet. The final dataset excludes frames with non-relevant labels and uses a face detector to identify and save the highest confidence face images to a folder.

In processing the AffectNet dataset, RetinaFace's flexibility was leveraged by adjusting its confidence score threshold to improve face detection rates. Initially set at a high threshold, resulting in only 16\% of faces detected, it was lowered to 0.4 and then to 0.1, increasing detection to 50\% but also raising the risk of false positives. This adjustment was critical for expanding the dataset to include a wider variety of facial features, which is crucial for training the GANmut model.

\begin{figure}[H]
    \includegraphics[width=1\columnwidth]{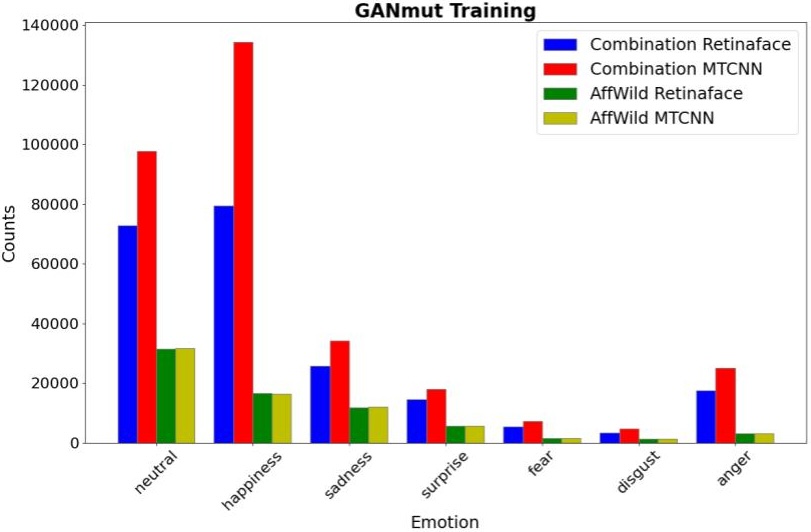}
    \caption{Emotion distribution for datasets used to train GANmut}
    \label{fig:emotions_distribution}
\end{figure}

\section{Numerical Results}

\subsection{VGGNet and ResNet}
The GANmut model uses metrics like Fréchet Emotion Distance and Smoothness Score to assess performance. These metrics compare features extracted from both the original dataset and the new dataset generated by GANmut using a classifier trained for emotion recognition. For classification, VGGNet \cite{simonyan2015deep} and ResNet50 \cite{resnet}, known for their deep learning efficiency, were adapted through fine-tuning for emotion classification, leveraging their pre-trained state on ImageNet dataset. Both models were adjusted to minimize overfitting and improve recognition accuracy in emotion classification tasks. Initial training on the RafDB dataset revealed issues of overfitting, with high training accuracy but poor validation performance. 

The final training configurations were optimized on the AffNet pre-cropped images, with balanced emotional categories to provide stable training outcomes and avoid abrupt accuracy changes. To improve performance and avoid overfitting, techniques like data augmentation, dropout layers, and learning rate schedulers were used. Subsequent training adjustments included unfreezing the last convolutional block to improve adaptability, which further increased validation accuracy. Both VGGNet and ResNet models were customized using additional dense layers and dropout to allow the model to learn more task-specific features.

Overall, changes to the classifiers and training methods have resulted in plateauing accuracy improvements, indicating that the model's potential has been fully utilized. Further exploration of image sizes and resolutions could potentially lead to enhanced outcomes.

\begin{table}[h]
    \centering
    \setlength{\arrayrulewidth}{0.3pt} 
    \captionsetup{font=normal, textfont=normal, format=plain, labelsep=colon, justification=centering}
    \begin{tabular}{|l|c|c|c|c|}
    \hline
    Model & Real & Combined & Aff-Wild2 & GANmut Original \\ \hline
    VGGNet   & 0.28 & 0.32 & 0.18 & 0.26 \\
    ResNet   & 0.47 & 0.44 & 0.18 & 0.33 \\ \hline
    \end{tabular}
    \caption{Average F1 Scores using RetinaFace (high values considered the best, max=1)}
    \label{tab:avg_f1_scores_retinaface}
\end{table}

\begin{table}[h]
    \centering
    \setlength{\arrayrulewidth}{0.3pt} 
    \captionsetup{font=normal, textfont=normal, format=plain, labelsep=colon, justification=centering}
    \begin{tabular}{|l|c|c|c|c|}
    \hline
    Model & Real & Combined & Aff-Wild2 & GANmut Original \\ \hline
    VGGNet   & 0.34 & 0.44 & 0.22 & 0.38 \\
    ResNet   & 0.50 & 0.58 & 0.27 & 0.48 \\ \hline
    \end{tabular}
    \caption{Average F1 Scores using MTCNN (high values considered the best, max=1)}
    \label{tab:avg_f1_scores_mtcnn}
\end{table}

\subsection{Fréchet Emotion Distance (FED)}
The FED score is a metric adapted from the Fréchet Inception Distance (FID) to assess the performance of GANs in generating emotionally expressive content. FED measures the similarity between the emotional features of generated and real images, using a classifier sensitive to emotional expressions to extract these features.

In experiments, the FED score reflects how closely the GAN-generated images match the emotional distribution of the original dataset. Features are typically extracted from either the last convolutional or dense layer of a classifier like VGGNet or ResNet. Lower FED scores indicate better performance, suggesting the generated content closely resembles the target emotional states.

Initial findings showed higher FED scores when using VGGNet compared to ResNet, suggesting that ResNet's deeper architecture with residual connections might capture emotional features more effectively. Various datasets and face detectors, including Aff-Wild2 with RetinaFace and MTCNN, were used to benchmark performance, indicating variations in effectiveness between the setups.

The experiments highlighted the influence of image quality and the classifier's training dataset on FED scores. Improvements were noted when switching to the RafDB dataset from the AffectNet small version, and ongoing adjustments to face detection and feature extraction methods continue to refine the GANmut model's accuracy in replicating emotional expressions.

\begin{table}[H]
    \centering
    \setlength{\arrayrulewidth}{0.3pt} 
    \captionsetup{font=normal, textfont=normal, format=plain, labelsep=colon, justification=centering}
    \begin{tabular}{|l|c|c|}
    \hline
    Model                  & VGGNet   & ResNet \\ \hline
    Aff-Wild2 Retinaface     & 35.96 & 8.0    \\
    Aff-Wild2 MTCNN          & 106.9 & 18.7   \\
    Combined Retinaface    & 31.34 & 6.5    \\
    Combined MTCNN         & 24.23 & 9.9    \\
    GANmut Retinaface      & 27.74 & 6.6    \\
    GANmut MTCNN           & 26.50 & 11.08  \\ \hline
    \end{tabular}
    \caption{FED scores (lower values considered the best showing that the distribution of generated images closely matches that of real images.)}
    \label{tab:fed_score}
\end{table}

\subsection{Smoothness Score}

The smoothness score is a metric that evaluates how naturally a model transitions between different emotional expressions, measuring the gradual changes in generated emotional expressions from a neutral state to a peak emotion across increasing intensities. It is calculated as the ratio between the maximum increase, which is the largest jump in confidence score between any two consecutive steps in a series of generated outputs, and the total variation range, which measures the difference in confidence scores from the most neutral to the most intense emotional expression. A lower ratio suggests more gradual transitions between emotional intensities, indicating smoother, more natural changes, while a higher ratio points to abrupt transitions, suggesting less realistic emotional evolution.

To compute this, an input image with a neutral expression undergoes a transformation through 10 interpolated steps for each basic emotion $c_i$. Each step increases in intensity, denoted by $\rho_j = 0.1 * j,\ j \in {1, 2, .., 10}$ and is assessed by a pre-trained classifier, which assigns a score to emotions at each intensity. The smoothness of the transition is determined as the ratio between the largest increase in these scores and the total variation across the series, with the results averaged across all emotions.

\begin{table}[h]
    \centering
    \setlength{\arrayrulewidth}{0.3pt} 
    \captionsetup{
        font=normal, 
        textfont=normal, 
        format=plain, 
        labelsep=colon, 
        justification=centering
    }
    \begin{tabular}{|l|c|c|}
    \hline
    Model                  & VGGNet   & ResNet \\ \hline
    Aff-Wild2 Retinaface     & 0.726 & 0.124    \\
    Aff-Wild2 MTCNN          & 0.165 & 0.643   \\
    Combined Retinaface    & 0.335 & 0.45    \\
    Combined MTCNN         & 0.376 & 0.382    \\
    GANmut Retinaface      & 0.427 & 0.38    \\
    GANmut MTCNN           & 0.054 & 0.36  \\ \hline
    \end{tabular}
    \caption{Smoothness Score (lower scores considered the best indicating a smooth progression)}
    \label{tab:performance_metrics}
\end{table}

\begin{figure}[H]
    \includegraphics[width=1\columnwidth]{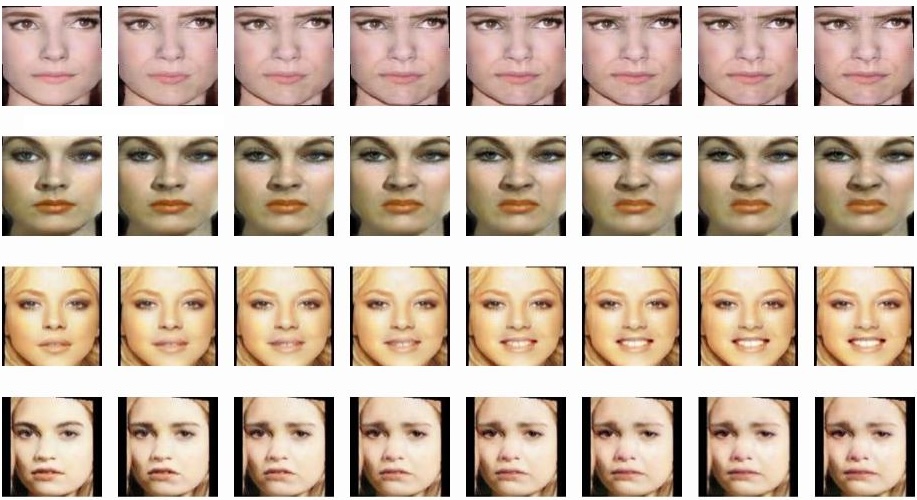}
    \caption{Increasing emotion-intensity; ordered top-bottom: anger, disgust, happiness, sadness.}
    \label{fig:smoothness_samples}
\end{figure}

\subsection{Discriminator F1}
The discriminator in a GAN acts as a binary classifier, determining whether input data is real (from the actual dataset) or fake (produced by the generator). Its effectiveness in distinguishing original from generated data enhances the generator's performance in creating more realistic outputs. To evaluate the discriminator, the F1 score is used, which is the harmonic mean of precision and recall. Precision measures the accuracy of fake data identified as fake, while recall assesses the correct identification of fake data against all actual fake data. An F1 score was computed using a custom dataset with generated images labeled as "fake" and original training images as "real," processed by the same data loader used during training.

\begin{table}[h]
    \centering
    \setlength{\arrayrulewidth}{0.3pt} 
    \captionsetup{
        font=normal,
        textfont=normal, 
        format=plain, 
        labelsep=colon, 
        justification=centering
    }
    \begin{tabular}{|p{0.9cm}|p{1.3cm}|p{1.3cm}|p{1.4cm}|p{1.4cm}|} 
    \hline
    Model & Aff-Wild2 Retinaface & Aff-Wild2 MTCNN & Combination Retinaface & Combination MTCNN \\ \hline 
    Average & 0.26 & 0.44 & 0.14 & 0.14 \\ \hline
    Real & 0.28 & 0.35 & 0.32 & 0.32 \\ \hline
    Fake & 0.37 & 0.55 & 0.09 & 0.09 \\ \hline
    \end{tabular}
    \caption{Discriminator F1 scores (lower values considered the best, indicating poor discrimination between real and generated images)}
    \label{tab:discriminator_f1}
\end{table}

\begin{figure}[h]
    \includegraphics[width=1\columnwidth]{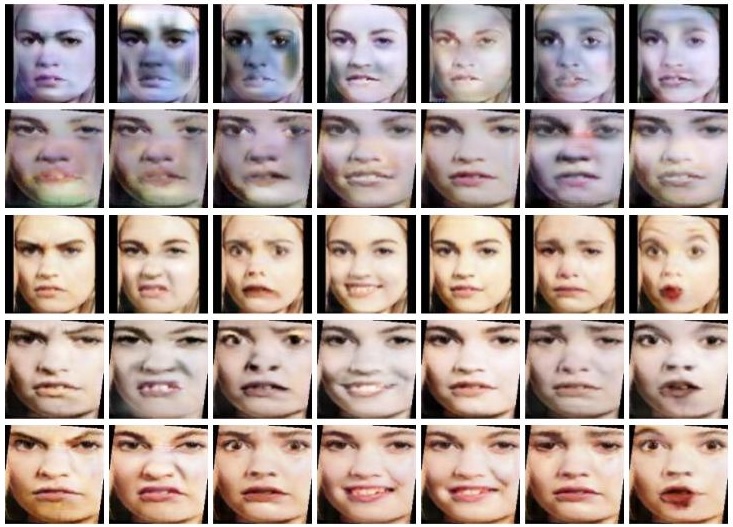}
    \caption{Top-bottom: Aff-Wild2 MTCNN; Aff-Wild2 RetinaFace; Combined MTCNN; Combined RetinaFace; GANmut RetinaFace. Left-Right: Anger; Disgust; Fear; Happiness; Neutral; Sadness; Surprise}
    \label{fig:visual_results}
\end{figure}

\section{Conclusion}
This study evaluated the GANmut model's performance through four benchmarking experiments using the Aff-Wild2 and a combined Aff-Wild2-AffNet dataset, alongside two face detectors, RetinaFace and MTCNN. Qualitatively, GANmut trained on the combined dataset showed visually superior results, capturing all seven target emotions and nuanced transitional expressions effectively. In contrast, models trained solely on Aff-Wild2 had difficulties in fully learning all emotions, leading to some inaccuracies. Quantitatively, the combined dataset models outperformed those trained only on Aff-Wild2, as evidenced by higher FED scores and Smoothness Scores. These metrics indicate better emotion classification accuracy and smoother transitions between emotions, suggesting a better representation of complex expressions like "fearfully surprised."

The improved performance on the combined dataset likely results from its larger size (about 270k images) and greater diversity, which provides a richer array of expressive features for the model to learn from. These results highlight the impact of dataset diversity and size on training effectiveness, particularly in "in-the-wild" conditions with variable head positions and lighting. In terms of face detection, models trained with RetinaFace generally yielded better FED scores, pointing to a slight advantage in using this detector for improved emotion recognition accuracy.

The study shows that the GANmut model, trained on a combined Aff-Wild2-AffNet dataset with both RetinaFace and MTCNN, more effectively captures complex emotions than training on Aff-Wild2 alone.

This research was supervised by Dr. Dimitrios Kollias, who provided invaluable support and guidance in shaping and executing the project. I am very grateful for his mentorship throughout the research process.

\begin{footnotesize}

\end{footnotesize}

\end{document}